\documentclass[letterpaper]{article} % DO NOT CHANGE THIS
\usepackage{aaai2026}  % DO NOT CHANGE THIS
\usepackage{times}  % DO NOT CHANGE THIS
\usepackage{helvet}  % DO NOT CHANGE THIS
\usepackage{courier}  % DO NOT CHANGE THIS
\usepackage[hyphens]{url}  % DO NOT CHANGE THIS
\usepackage{graphicx} % DO NOT CHANGE THIS
\usepackage{natbib}  % DO NOT CHANGE THIS AND DO NOT ADD ANY OPTIONS TO IT
\usepackage{caption} % DO NOT CHANGE THIS AND DO NOT ADD ANY OPTIONS TO IT
\usepackage{multirow}
\usepackage{amsmath}
\usepackage{amssymb}

\usepackage{bibentry}
\newcommand{\mathbold}[1]{\ensuremath{\boldsymbol{\mathbf{#1}}}}

\newcommand{\nestedmathbold}[1]{{\mathbold{#1}}}

\newcommand{\mbv}{\nestedmathbold{v}}

\newcommand{\mbx}{\nestedmathbold{x}}
\newcommand{\mby}{\nestedmathbold{y}}

\newcommand{\mbI}{\nestedmathbold{I}}

\newcommand{\mbV}{\nestedmathbold{V}}

\newcommand{\mbepsilon}{\nestedmathbold{\epsilon}}

\newcommand{\mbtheta}{\nestedmathbold{\theta}}

\newcommand{\mbzero}{\nestedmathbold{0}}

\newcommand{\cL}{\mathcal{L}}
\newcommand{\cN}{\mathcal{N}}

\usepackage{booktabs}
\usepackage{algorithm}
\usepackage{algpseudocode}
\usepackage{xcolor}  
\usepackage{sidecap}

\usepackage{afterpage}
\usepackage{caption}
\usepackage{cuted}

\usepackage{newfloat}
\usepackage{listings}
\DeclareCaptionStyle{ruled}{labelfont=normalfont,labelsep=colon,strut=off} % DO NOT CHANGE THIS
\floatstyle{ruled}
\newfloat{listing}{tb}{lst}{}
\floatname{listing}{Listing}
\nocopyright 
\title{Nested AutoRegressive Models}
\author{
    Hongyu Wu,
    Xuhui Fan\footnote{Corresponding author, xuhui.fan@mq.edu.au},
    Zhangkai Wu,
    Longbing Cao
}
\affiliations{
    \textsuperscript{\rm 1} Macquarie University, Australia
}

\usepackage{float}

\usepackage[capitalize]{cleveref}
\begin{document}

\maketitle

\begin{abstract}
AutoRegressive (AR) models have demonstrated competitive performance in image generation, achieving results comparable to those of diffusion models. However, their token-by-token image generation mechanism remains computationally intensive and existing solutions such as VAR often lead to limited sample diversity. In this work, we propose a Nested AutoRegressive~(NestAR) model, which proposes nested AutoRegressive architectures in generating images. NestAR designs multi-scale modules in a hierarchical order. These different scaled modules are constructed in an AR architecture, where one larger-scale module is conditioned on outputs from its previous smaller-scale module. Within each module, NestAR uses another AR structure to generate ``patches'' of tokens. The proposed nested AR architecture reduces the overall complexity from $\mathcal{O}(n)$ to $\mathcal{O}(\log n)$ in generating $n$ image tokens, as well as increases image diversities. NestAR  further incorporates flow matching loss to use continuous tokens, and develops objectives to coordinate these multi-scale modules in model training. NestAR achieves competitive image generation performance while significantly lowering computational cost.
\end{abstract}


\section{Introduction}
AutoRegressive (AR) models, built on the next-token prediction paradigm, form the foundation of large language models (LLMs), aligning naturally with the next-token (or word) prediction task~\cite{llama2,gpt4,attn}. While AR models have long been central to natural language processing, recent studies have reintroduced them as strong competitors~\cite{var,llamagen} to the widely used diffusion models~\cite{diff_m,latent_dm} in the field of image generation.

% Following the success of AR models in natural language processing (NLP), pioneer researchers have adapted AR models  to image generation with SOTA results surpassing other popular models such as diffusion model (DM) . 

% \begin{figure}[htbp]
%   \centering
%   \begin{minipage}[c]{0.40\textwidth}
%   \raggedright
%     \includegraphics[trim=150 600 20 60, clip, width=\linewidth]{images/NestAR_fig1_2.pdf}
%   \end{minipage}%
%   \hspace{-40pt}
%   \begin{minipage}[c]{0.10\textwidth}
%     \raggedleft
%     \captionof{figure}{\small }
%     \label{Fig1}
%   \end{minipage}
% \end{figure}

\begin{figure}[ht]
  \centering
    \noindent
    \includegraphics[width=0.4\linewidth]{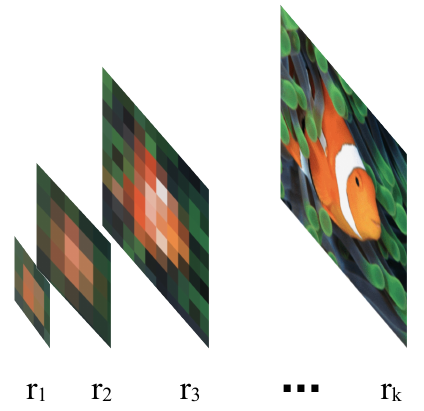}
    \hfill
    \includegraphics[width=0.4\linewidth]{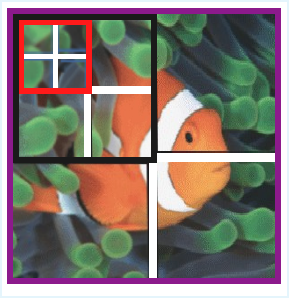}
    \caption{Visual comparison between VAR (left panel) and the proposed NestAR (right panel). VAR: next resolution prediction from coarse to fine resolutions of the entire image. A single AR model all $K$ resolutions. (b) NestAR with $3$ scale modules: different scaled modules generating progressive larger area of the image. These modules are bounded by red, black, and purple boxes correspondingly.}
      \label{Fig1}
\end{figure}

Recent studies on image generation using AR models not only achieve state-of-the-art~(SOTA) results, but also open up new research directions. LlamaGen \cite{llamagen} successfully adopts the Llama architecture to image generation. LlamaGen outperforms popular diffusion models and significantly speeds up inference time, making it become the foundation for many models \cite{var, next_patch}. MAR \cite{mar} was one of the first work to introduce continuous tokens to AR models, pushing the boundaries of the capabilities of AR models and allowing variants such as Fluid \cite{fluid}, xAR \cite{xar} to achieve SOTA results.

However, due to its token-by-token generation nature, AR models are usually concerned about requiring a long running time in image generation. Existing methods~\cite{var,distil,xar,next_patch} trying to address this issue either sacrifices image quality (higher FID) or diversity (lower IS)~\cite{arsurvey,xar,next_patch,distil}. Thus, developing approaches that can retain both quality and diversity while improving speed remains an open research problem.

In this paper, we propose the Nested AutoRegressive~(NestAR) model, for fast generating images as well as increasing generation diversities. NestAR designs a nested AutoRegressive architecture that consists of two levels of AR structures to generate images. The first level is a hierarchical multi-scale architecture, in which each scaled module is responsible to generate a scale-specific number of image tokens. More importantly, the current scaled module depends on the outputs from previous scaled modules. Within each scaled module, the second level is another AR structure that generates ``patches'' of tokens conditioned on previously generated patches of tokens, including those generated by previous modules.

By setting patch sizes to increase exponentially with the module number, NestAR takes fewer steps $\mathcal{O}(\log(n))$ than token-by-token generation approach $\mathcal{O}(n)$ at inference time. At the same time, the AR structure within each scaled module enables to generate more diversified images. See Figure~\ref{Fig1} for visualizing a 3-module NestAR. 

Regarding the conditional distribution, NestAR uses a flow matching mechanism to conditionally generate the next patch of tokens, which uses continuous tokens to preserve token information. Additionally, we introduce an objective that compares the velocities of different modules on the same image, effectively coordinating their behavior throughout the training process.

The main contributions of NestAR can be summarized as:
\begin{itemize}
\item Using a hierarchical architecture, NestAR reduces the computational complexity of generating images from $n$ to $\log(n)$.
\item By adopting the AR architecture within each scaled module, NestAR increases the diversity of image generation.
\item We design an objective that coordinate different modules' behaviour during the training process. 
%\item 
\end{itemize}
Extensive experimental evaluations {show NestAR achieves a new highest IS score; its generation speed beats most diffusion based and AR based models; while maintaining competitive FID score.}

\section{Preliminaries}
\subsection{AutoRegressive Model}
AR models refer to one of the fundamental generative models that are popularly used for tasks such as image and audio generation. They sequentially operate by generating each token, such as a word, pixel or feature, with each step conditioned on the elements produced in previous steps. Formally, given a random variable $\mbx$ arranged in a sequence of $n$ tokens $(\mbx_1,\mbx_2, ...,\mbx_n)$, the AR model learns the conditional distribution of a particular token, given all previous tokens:
\begin{align}
p(\mbx_i | \mbx_{<i}) = p(\mbx_i|\mbx_1, \mbx_2, ..., \mbx_{i-1}), 
\end{align}
and the joint distribution $p(\mbx_1, \ldots, \mbx_n)$ is given as $p(\mbx) = \prod_{i=1}^{n}p(\mbx_i|\mbx_{<i})$. At the inference time, an AR model samples tokens one by one using the learned conditional distribution for each token.

\subsection{AutoRegressive Model for Images}
Inspired by the success of AR models in NLP, early efforts in adopting AR models to image 
generation focused on quantize two dimensional images into a sequence of discrete 
tokens. VQVAE \cite{vqvae} uses a Variational autoencoder to map an image to a feature map then 
uses a quantiser to build a codebook which converts the feature map into a one-dimensional 
discrete tokens. This process is reversed to re-construct the original image. VQGAN 
\cite{vqgan} enhances this process by adding adversarial loss. LlamaGen \cite{llamagen} uses Llama's LLM architecture to implement an AR model to generate images, which achieves SOTA results while unifying language and image generation models. However, the model uses 256 steps to generate each image. \cite{next_patch} uses the discrete feature map from LlamaGen to form patches as tokens. It saves significant computational resource but is still trained on all individual tokens over the entire training cycle. 

The study in \cite{mar} represents one of the first to use continuous tokens, creating an AR model that uses 
diffusion loss instead of the traditional cross entropy loss. xAR \cite{xar} uses the continuous feature maps from \cite{mar} to form patches for reducing the generation steps successfully, while sacrificing image diversity.

Some recent work explored the design of tokens to improve the generation speed and quality 
of the results. \cite{wavelet} uses wavelets, \cite{var} uses resolution as tokens, all achieving good results. \cite{distil} is the first to distill from an AutoRegressive model to generate an image in one or few steps with respectable results.

\subsection{Diffusion and Flow Matching Models}
Diffusion and Flow matching models are popular generative models for image generation. They are capable of transforming sampled noise into crystal clear images through either a multi-step denoising process (DM) \cite{diff_m,latent_dm, consist_m} or a direct mapping between the data distribution to a standard Gaussian distribution (FM) \cite{flow, rect_flow}.

\section{Methodology}
In this section, we first introduce and formulate the generative process of the NestAR model. Then, we detail the architectural design of NestAR, followed by a discussion on the training and sampling methods.

\subsection{Nested AutoRegressive Model}
In most AutoRegressive~(AR) approaches, only one single AR model is trained to generate all  $n$ tokens. This token-by-token sequential generation nature is time consuming, as the complexity scales to $\mathcal{O}(n)$. In the proposed NestAR, a series of differently scaled modules are stacked in a hierarchical architecture. Up to $M=\log_k(n)$ differently scaled modules are developed, where $k$ is the number of evaluations for modules to generate tokens.

Given a sequence of $n$ tokens $\{\mbx_{1}, \mbx_{2},\ldots, \mbx_{n}\}$ representing an image or a feature map of an image, these tokens are ordered based on a particular scanning method, such as roster scan~\cite{raster}. We denote $\mbx_{m:n}=\{\mbx_m, \mbx_{m+1}, \ldots, \mbx_n\} (m<n)$ and $\mbx_{m,i}=\mbx_{(k^{m-1}\cdot(i-1)+1):k^{m-1}\cdot i}$ for notational convenience, with $\mbx_{m,i}$ also representing the $i$-th patch in the $m$-th scaled module.

NestAR assumes each scaled module generates the ``patches'' of tokens, rather than individual tokens, per one evaluation. That is, evaluating the $m$-th~($1\le m\le M$) module would generate a patch which comprises $k^{m-1}$ tokens. As each module is evaluated for $k$ times, a total of $k\cdot k^{m-1}=k^m$ tokens are generated.

\begin{figure*}[ht]
    \includegraphics[trim=44 270 150 180, clip, width=\linewidth]{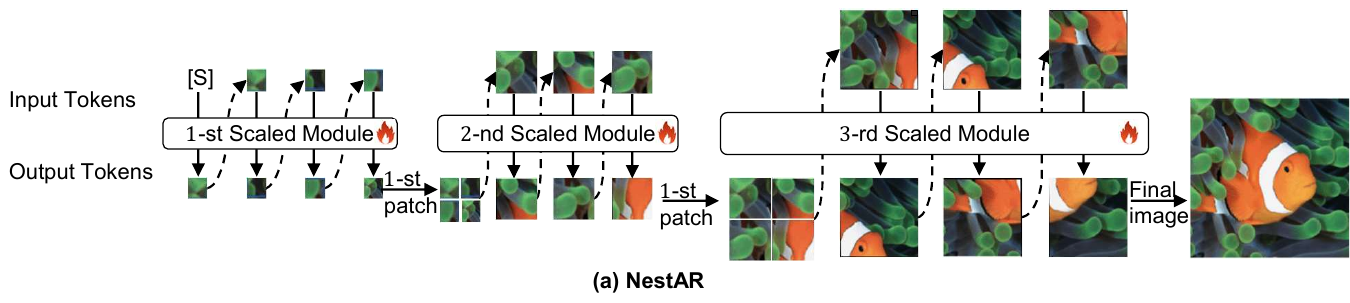}
    \includegraphics[trim=60 210 100 200, clip, width=\linewidth]{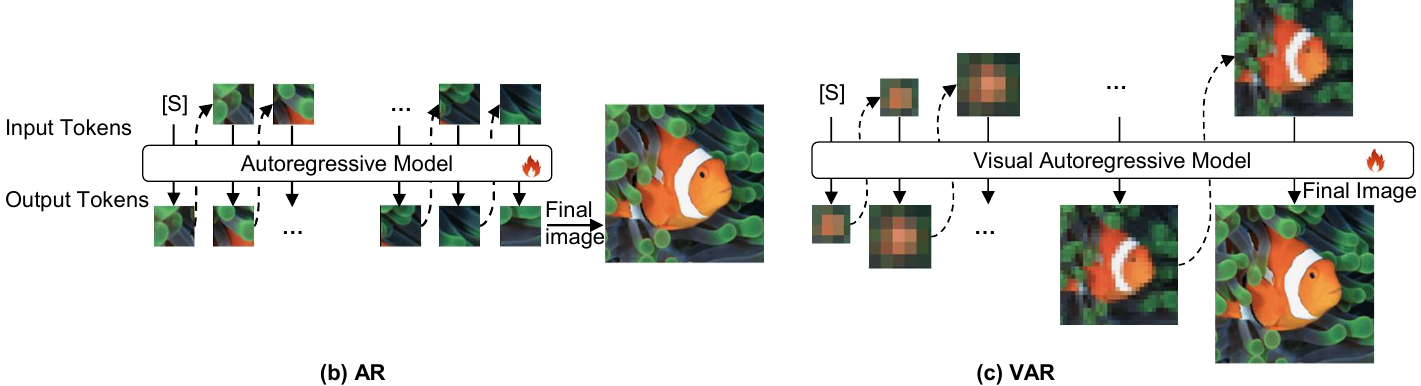}
    % \includegraphics[trim=300 35 0 0, clip, width=\linewidth]{images/hear_fig2_trim.pdf}
    % \includegraphics[trim=0 35 550 0, clip, width=0.45\linewidth]{images/hear_fig2_trim.pdf}
    % \includegraphics[trim=170 260 300 240, clip, width=0.5\linewidth]{images/AR-gm.pdf}
    % \hfill
    % \includegraphics[trim=170 180 250 220, clip, width=0.5\linewidth]{images/VAR-gm.pdf}
    % \includegraphics[width=0.45\linewidth]{images/var_gen_trim.pdf}
    \caption{Visualization of the mechanisms for different AR models. (a), NestAR model; (b), vanilla AR model; (c) Visual AutoRegressive Model~(VAR). NestAR expands the sizes of the patches along with the module orders. The $1$-st scaled module, which is also the smallest, captures the distribution of the smallest patch of tokens. Its output then becomes the first patch of tokens for the $2$-nd scaled module which models a larger-sized patch of tokens. This process continues to the highest scaled-module, in which the generated patches can form the entire image. Vanilla AR generates tokens one at a time based on previous tokens. VAR generates different resolutions of the entire image in a hierarchical manner.}
    \centering
    \label{Fig2.}
\end{figure*}

To begin the detailed procedure, the $1$-st scaled module models $k$ tokens through a conventional AR architecture as:
\begin{align}
P(\mbx_{1:k}) = \prod_{i=1}^{k}P_{\mbtheta_1}(\mbx_i|\mbx_{1:i-1}),
\end{align}
where $P_{\mbtheta_1}(\mbx_i|\mbx_{1:i-1})$ defines the conditional distribution of $\mbx_i$ given its previous tokens and $\mbtheta_1$ is the corresponding parameter.

Given the tokens $\mbx_{1:k^{m-1}}$ generated from all the $(m-1)$ ($1<m\leq M$) scaled modules, the $m$-th  scaled module aims to model tokens $\mbx_{(k^{m-1}+1):k^m}$, with the AR architecture in the conditional distribution over the tokens $\mbx_{(k^{m-1}+1):k^{m}}$ designed as:
\begin{multline} \label{eq:conditional-distribution-in-mth-module}
P(\mbx_{(k^{m-1}+1):k^{m}}|\mbx_{1:k^{m-1}}) \\
= \prod_{i=2}^{k}P_{\mbtheta_m}(\mbx_{m,i}|\mbx_{1:k^{m-1}\cdot(i-1)}),    
\end{multline}
where $P_{\mbtheta_m}(\mbx_{m,i}|\mbx_{1:k^{m-1}\cdot(i-1)})$ defines the conditional distribution of $\mbx_{m,i}$ given the historical tokens. These historical tokens include $\mbx_{1:k^{m-1}}$ from all the previous modules and $\mbx_{(k^{m-1}+1):k^{m-1}\cdot (i-1)}$ from all the previous $(i-1)$ evaluations in the same scaled $m$-th module; and where $\mbtheta_m$ is the corresponding parameter. As can be seen in \cref{eq:conditional-distribution-in-mth-module}, each $m$-th ($1<m\leq M$) scaled module generates patches from the $2$-nd position in each layer, whereas the $1$-st patch is already formed by previous layers as $\mbx_{1:k^{m-1}}$. As a result, it 
would be evaluated for $k-1$ times to generate the tokens $\mbx_{(k^{m-1}+1):k^{m}}$.

Given such constructions, the joint distribution $P(\mbx_{1:k^{M}})$ can be written as:
\begin{multline} \label{eq:full-joint-distribution-over-x}
P(\mbx_{1:k^{M}})
=\prod_{m=1}^{M}\prod_{i=2}^{k}P_{\mbtheta_m}(\mbx_{m,i}|\mbx_{1:k^{m-1}\cdot(i-1)}). 
\end{multline}
That is, while \cref{eq:full-joint-distribution-over-x} decomposes the joint distribution over all tokens, it does so without imposing any structural assumptions. Figure~\ref{Fig2.} gives a visualization of vanilla AR and NestAR over the token generation. 

\subsubsection{Two Levels of AR Structures in NestAR.} Two AR structures are used to decompose the joint distribution in \cref{eq:full-joint-distribution-over-x}: (1) each scaled $m$-th module is proceeded conditioned on patches of tokens from previous modules including $\mbx_{1:k}, \mbx_{(k+1):k^2}, \ldots, \mbx_{(k^{m-2}+1):k^{m-1}}$; (2) the patches of tokens within each scaled $m$-th module are proceeded in an AR way.  

On one hand, the scale-wise AR structure enables batch generations of tokens. In particular, the proposed NestAR only evaluates the conditional distribution for $(k-1)\cdot m+1$ times to generate $k^m$ tokens, which is $\log$ scale of the conventional AR models. On the other hand, the patch-wise AR structure within each scale allows more diverse image generation.

% \hongyu{Each AR model models tokens 1 to $k^m$. The first $k^{m-1}$ tokens are modeled both by the $(m-1)^{th}$ model and the first token of the $m^{th}$ model, hence the $m^{th}$ model can be written using the $(m-1)^{th}$} \xuhui{In generation, which one do you use?}\hongyu{$(m-1)^{th}$ in generation}

\subsection{Calculating the Conditional Distribution}
The commonly-used ``discrete tokens” may result in information loss~\cite{disInfoLoss}, or inconsistencies in codebook construction~\cite{codebookissue1, codebookissue2}. Instead, NestAR adopts the continuous token approach~\cite{mar}, which uses the flow matching~\cite{flow, rect_flow} mechanism to generate patches of tokens conditioned on previously generated tokens. 

For the $i$-th evaluation in the $m$-th scaled module, the flow matching approach aims to generate a patch of tokens $\mbx_{m,i}$, conditioned on the generated tokens $\mbx_{1:k^{m-1}\cdot (i-1)}$. To proceed with the training objective, a white noise sample $\mbepsilon_{m,i}$, which shares the same size as $\mbx_{m,i}$, is first sampled from $\cN(\mbzero, \mbI)$. Let the time step $t\sim\text{Uniform}[0, 1]$, the interpolation input $\mby_{t}$ is calculated as $\mby_{t} = (1-t)\mbx_{m,i} + t \mbepsilon_{m,i}$. 

When $\mbx_{m,i}$ is known, the ground-truth velocity of $\mby_t$ at the time step $t$ can be calculated as $\mathrm{d}\mby_{t}/\mathrm{d}t=\mbepsilon_{m,i}-\mbx_{m,i}$. Since $\mbx_{m,i}$ is unknown in practice, a velocity approximation $\mbv_{\mbtheta_m}(\mby_t, t| \mbx_{1:k^{m-1}\cdot (i-1)})$ is developed, conditioned on the generated tokens $\mbx_{1:k^{m-1}\cdot (i-1)}$. The training target is to approximate $\mbv_{\mbtheta_m}(\mby_t, t| \mbx_{1:k^{m-1}\cdot (i-1)})$ to the ground-truth velocity $\mbepsilon_{m,i} - \mbx_{m,i}$ as:
\begin{multline} \label{eq:flow-matching-loss}
\cL_{\text{module},m}=\mathbb{E}_{t, i}\left[\|\mbv_{\mbtheta_m}(\mby_t, t| \mbx_{1:k^{m-1}\cdot (i-1)})\right.\\
\left.-(\mbepsilon_{m,i} - \mbx_{m,i})\|^2\right]. 
\end{multline} 
Given the image data, it is noted that every $m$-th scaled module modeled parts of the image, i.e. the first $k^{m}$  tokens only.

Comparing with the cross-entropy loss used in common AutoRegressive models with discrete tokens, continuous tokens offer a more accurate representation of images. Additionally, flow matching models provide greater expressive power for modeling complex probability distributions. These advantages make the flow-matching AutoRegressive approach a compelling choice for the NestAR framework.

% \xuhui{is there something wrong with the following equation? I've updated to the above one. Please check!}
% $f_{\theta}$ is the AutoRegressive model with parameter theta trained to predict $\mbx_q$, by optimising the velocity at $t_p$
% \begin{align} \label{eq:loss}
% L=\sum_{q=1}^{k}\|f_{\theta}(\{Y_1^{t_p}...Y_q^{t_p}\},t_p )- (\epsilon_q - \mbx_q)\|^2\
% \end{align}

\subsection{Coordinating Scaled Modules}
While individual scaled modules can be trained independently, introducing an objective which coordinates different scaled modules might be important to improve the overall performance. The simplest coordinating strategy would be calculating and then maximizing the log-likelihood over all tokens. Since we estimates velocities for patches of tokens, the instantaneous change of variables theorem~\cite{NeuralODE, flow} can be leveraged to compute this log-likelihood. However, such method is computationally intensive~\cite{ODEIntensive} and may not be practical for large datasets.

As an alternative, we choose to compare velocities of consecutive modules to coordinate different scaled modules. 
That is, the distribution of the $1$-st patch in the $m$-th ($1<m\le M$) module should match to those of patches generated from previous modules. Let $\widetilde{\mbV}_{m-1}=[\mbv_{\mbtheta_{m-1}}(\mby_{m-1,i},t), \ldots, \mbv_{\mbtheta_{m-1}}(\mby_{m-1, k},t)]^{\top}$ denotes the concatenation of $K$ velocities $\{\mbv_{\mbtheta_{m-1}}(\mby_{m-1,i},t)\}_i$, $\widetilde{\mbV}_{m-1}$ is expected to match the velocity of the same patch in the $m$-th module. As a result, the coordinating objective can be written as:
\begin{align} \label{eq:tune}
\cL_{\text{coor},m}=\mathbb{E}_{t}\|\widetilde{\mbV}_{m-1}- \mbv_{\mbtheta_{m}}(\mby_{t},t|\mbx_{1:k^{m-1}})\|^2,
\end{align}
where $t$ follows the time step distribution. 

% Where $f_{\theta_m}$ and $f_{\theta_{m+1}}$ are the pre-trained $m^{th}$ and $(m+1)^{th}$ scaled modules, t is sampled from $U[0,1]$, $Y_m$ is the interpolated input at time t to model $f_{\theta_m}$ and $Y_m$ = $Y_{m+1}[1]$, i.e. the first token of model $f_{\theta_{m+1}}$'s interpolated input $Y_{m+1}$, $f_{\theta_{m+1}}(Y_{m+1},t)[1]$ is the velocity of $f_{\theta_{m+1}}$'s first token.

The objective function of NestAR would be:
\begin{align}
    \cL = \lambda_{\text{module}}\sum_{m=1}^M \cL_{\text{module},m} + \lambda_{\text{coor}}\sum_{m=1}^M\cL_{\text{coor},m},
\end{align}
where $\lambda_{\text{module}}$ and $\lambda_{\text{coor}}$ refer to the corresponding coefficients. 

\subsection{Image Generation in NestAR}
After training the velocities $\{\mbv_{\mbtheta_m}(\cdots)\}_{m}$ in NestAR, images can be generated following its generative process. Particularly, each $i$-th patch in the $m$-th module can be generated through an ordinary differential equation~(ODE) as $\mbx_{m,i}=\mbepsilon_{m,i}+\int_{1}^0\mbv_{\mbtheta_m}(\mby_{t}, t|\mbx_{1:k^{m-1}\cdot (i-1)})\textrm{d}t, \forall 2\le i\le k, 1\le m\le M$, which can be approximated by an ODE-solver such as Euler approximation~\cite{flow}. We write it as $\mbx_{m,i}=\text{ODE-solver}(\mbv_{\mbtheta_m}, \mbepsilon_{m,i}, \mbx_{1:k^{m-1}\cdot (i-1)})$.

\cref{alg:sample} shows the detailed procedures. Lines $1\sim 2$ first generate the first token $\mbx_1$. For $m\in\{1, \ldots, M\}$ and  $i\in \{2, \ldots, k\}$, an ODE-solver generates the token for the $i$-th patch in the $m$-th module. Noted that, except for the $1$-st module, patches are generated from the $2$nd to the $k$-th, since the $1$st patch has been generated through the previous scaled modules already. All the generated patches of tokens can be concatenated to form the whole set of tokens for the image. 
\begin{algorithm}[h]
\caption{Image generation in NestAR}\label{alg:sample}
\begin{algorithmic}[1]
    \Require Trained velocities $\mbv_{\mbtheta_1}(\ldots), \ldots, \mbv_{\mbtheta_M}(\ldots)$; $M$: number of scaled modules $M$; $k$: number of patches in each module
    \Ensure $\mbx_{1:k^M}=\mbx_{1:n}$
    \State Sample $\mbepsilon_1\sim \cN(\mbzero, \mbI)$ 
    % \Comment{$\mbepsilon_1$ shares the same shape of $\mbx_1$}
    \State $\mbx_1=\text{ODE-solver}(\mbv_{\mbtheta_1},\mbepsilon_1, \emptyset))$ 
    % \Comment{use ODE-solver to generate $\mbx_1$}
    \For {$m = 1, 2, \ldots, M$} \Comment{each $m$-th module}
        \For{$i = 2, \ldots, k$} \Comment{each $i$-th patch}
        \State Sample $\mbepsilon_{m,i}\sim \cN(\mbzero, \mbI)$ 
        % \Comment{$\mbepsilon_{m,i}$ shares the same shape of $\mbx_{1:k^{m-1}}$}
        \State $\mbx_{m,i}=\text{ODE-solver}(\mbv_{\mbtheta_m}, \mbepsilon_{m,i}, \mbx_{1:k^{m-1}\cdot (i-1)})$ 
        % \Comment{use ODE-solver to generate $\mbx_{m,i}$}
        \EndFor
    \EndFor
    \State \Return $\mbx_{1:k^M}$
\end{algorithmic}
\end{algorithm}
\begin{table*}[h]
\centering
%\resizebox{1\textwidth}{!}{
\begin{tabular}{ccccccc}
\toprule
% \cmidrule(lr){3-5} \cmidrule(lr){6-7}
\textbf{Type} & \textbf{Model} & \textbf{Params}  & \textbf{FID}$\downarrow$ & \textbf{IS}$\uparrow$ & \textbf{Precision}$\uparrow$ & \textbf{Recall}$\uparrow$  \\
\midrule
GAN&GigaGAN\cite{gigaGan}&569M&3.45&225.5&0.84&0.61\\
GAN&StyleGan-XL\cite{styleGan}&166M&2.3&265.1&0.78&0.53\\
\midrule 
Diffusion&ADM\cite{adm}&554M&10.94&101.2&0.69&0.63\\
Diffusion&LDM-4\cite{latent_dm}&400M&3.6&247.7&0.87&0.48\\
Diffusion&DiT-XL/2\cite{dit}&675M&2.27&278.2&0.83&0.57\\
\midrule 
Flow-Matching&SiT-XL/2\cite{sit}&675M&2.06&277.5&0.83&0.59\\
Flow-Matching&REPA\cite{repa}&675M&1.8&284&0.81&0.61\\
\midrule 
Mask.&MaskGIT\cite{maskgit}&227M&6.18&182.1&0.8&0.51\\
Mask.&MAGVIT-v2\cite{maskvit}&307M&1.78&319.4&--&--\\
\midrule 
AR&VQGAN\cite{vqgan}&227M&18.65&80.4&0.78&0.26\\
AR&VQGAN\cite{vqgan}&1.4B&15.78&74.3&--&--\\
AR&ViTVQ\cite{vit}&1.7B&4.17&175.1&-&-\\
AR&DART-AR\cite{dart}&812M&3.98&256.8&-&-\\
AR&MonoFormer\cite{mono}&1.1B&2.57&272.6&0.84&0.56\\
AR&LlamaGen-3B\cite{llamagen}&3.1B&2.18&263.3&0.81&0.58\\
AR&LlamaGen-L\cite{llamagen}&343M&3.07&256.06&0.83&0.52\\
\midrule 
MAR&MAR-B\cite{mar}&208M&2.31&281.7&0.82&0.57\\
MAR&MAR-L\cite{mar}&479M&1.78&296&0.81&0.6\\
MAR&MAR-H\cite{mar}&943M&1.55&303.7&0.81&0.62\\
\midrule 
VAR&VAR-$d$16\cite{var}&310M&3.3&274.4&0.84&0.51\\
VAR&VAR-$d$\cite{var}20&600M&2.57&302.6&0.83&0.56\\
VAR&VAR-$d$\cite{var}30&2.0B&1.97&323.1&0.82&0.59\\
\midrule
xAR&XAR-B\cite{xar}&172M&1.72&280.4&0.82&0.59\\
xAR&XAR-L\cite{xar}&608M&1.28&292.5&0.82&0.62\\
xAR&XAR-H\cite{xar}&1.1B&1.24&301.6&0.83&0.64\\
\midrule 
NestAR&NestAR-B&344M&2.86&320.6&0.54&0.78\\
NestAR&NestAR-L&780M&2.29&338.3&0.57&0.78\\
NestAR&NestAR-H&1.3B&2.22&342.4&0.57&0.79\\ 
\bottomrule
\end{tabular}
\captionof{table}{Comparison of generation performance on ImageNet-256. Metrics include Fr\'{e}chet Inception Distance (FID$\downarrow$), Inception Score (IS$\uparrow$), Precision$\uparrow$, and Recall$\uparrow$.}
\label{tab:main}
%\end{center}
\end{table*}

% \begin{figure}[htbp]
%     % \includegraphics[trim=0 30 0 300, clip, scale=0.58]{images/HeAR_fig2_3.pdf}
%     \includegraphics[trim=0 35 550 0, clip, width=0.45\linewidth]{images/hear_fig2_trim.pdf}
%     \caption{Vanilla AR and VAR.}
%     \centering
%     \label{fig:related-work}
% \end{figure}

\subsection{Connections to Existing AR Models}
The proposed NestAR model shares deep connections with existing autoregressive (AR) models but also introduces significant differences. Its hierarchical multi-scale modules function similarly to VAR, yet they generate only sub-regions of images, unlike VAR which typically generates different resolutions of the entire image. NestAR also follows MAR in using the flow matching mechanism to generate samples from conditional distributions. Furthermore, NestAR's patch generation, rather than token generation, is similar to xAR. However, while xAR uses only one module to generate same size patches, NestAR develops Nested AR structures that involve multiple multi-scale modules to generate patches in different sizes.

% \vspace{-10pt}
% \begin{strip}
%\clearpage
%\onecolumn

%\begin{center}
%\begin{table*}[!htbp]
%}
%\twocolumn
% \end{strip}

\begin{figure*}[h]
\centering
\resizebox{\textwidth}{!}{
\includegraphics{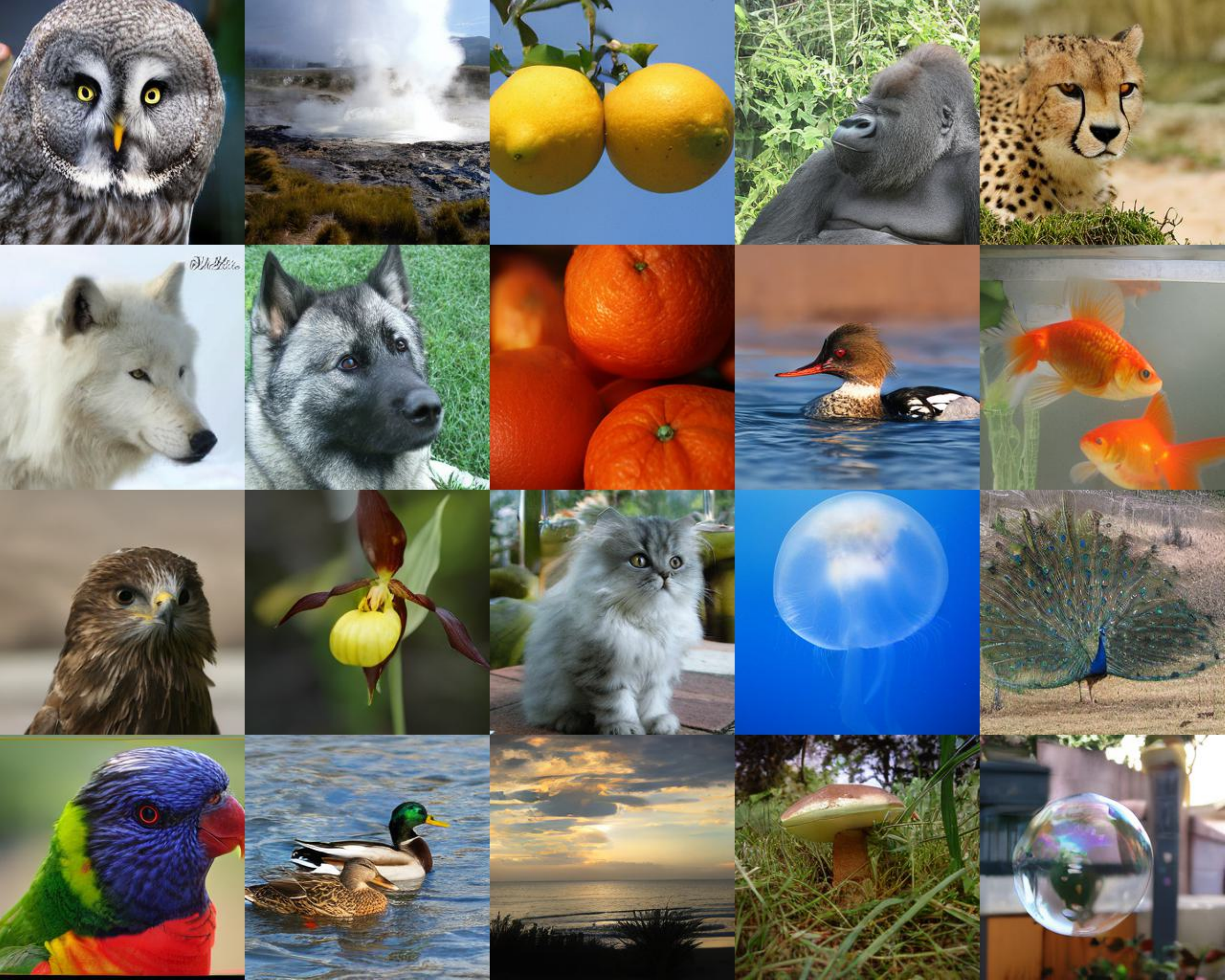}}
\captionof{figure}{Qualitative Results: Generated $256\times 256$ image samples from our NestAR-H model.}
\label{image1}
\end{figure*}

\section{Experiments}
We test the performance of NestAR on the ImageNet~\cite{imagenet} dataset at $256\times 256$ resolution. In order to comprehensively evaluate its generation capability, we set up several tasks to asnswer the following questions:
\begin{description}
    \item[RQ1:] How does NestAR perform when compared with state-of-the-art methods in geneated image qualities? 
    % \hongyu{by SOTA, do you mean xAR? Below is my explanation, not sure how much and how best to phrase it in the paper.}
    % \hongyu{In an ideal situation, each scaled module could perfectly learn the probability distribution of its feature map. In this case, the probability distribution of the $1^{st}$ scaled module would exactly match the probability distribution of the $1^{st}$ token of the $2^{nd}$ scaled module. In theory, the modules could be connected to work coherently to generate images. However, in practice, the models could only approximate the probability distribution of its target. Hence the the $1^{st}$ scaled module and the $1^{st}$ token of the $2^{nd}$ scaled module have different approximations of the same target. Given sufficient resources, the scaled modules could be trained together from scratch to work coherently. However, given the resource and time constraints, to demonstrate the theory of the Nested AutoRegressive model, pre-trained $2^{nd}$ scaled module was adopted, and fine-tuning is used to match the $1^{st}$ scaled module with the $1^{st}$ token of the $2^{nd}$ scaled module. As a result the performance is bounded by the pre-trained $2^{nd}$ scaled module.}
    \item[RQ2:] How does NestAR compare with other methods in terms of generation speed? 
    \item[RQ3:] How does the size of $1$-st scaled module would affect the model performance? 
\end{description}

\subsection{Experimental Settings} \label{implement}
\subsubsection{Encoder and Decoder.} In loading ImageNet, we use the public available tokenizer KL-$16$ provided by LDM \cite{latent_dm} (instead of VQ-VAE to avoid quantization loss). The tokenizer uses a downsampling scale~$r=16$~\cite{mar} to convert an image to a continuous latent representation $I=R(h=16, w=16, c=3)$, where $h,w,c$ are the height, width and number of channels for the representation. Regarding generating images, the generated latent representation will be passed through the decoder in LDM to produce the image.

\subsubsection{Algorithm Settings.} Following \cite{xar}, we use the raster token order~\cite{raster} to arrange the tokens sequentially. That is, the tokens are ordered as starting from top left and going left to right and top to bottom. The number of evaluations within each module is chosen as $k=4$, through which the patches within each module can be formed as a square.  
% Regarding the number of tokens in each AR level, $k$ is chosen as $k=4$. , each token is a square patch of the feature map. 

\noindent\textbf{Hyper-parameter settings} The pre-trained AR models are combined to form the NestAR model. The hyper-parameters used for tuning are detailed in the appendix. Each scaled AR is pre-trained using the same hyper-parameters as in \cite{dit,mar}. Table~\ref{tab:size} contains the parameter size for each scaled AR.

\begin{table}[t]
%\begin{table}[!htbp]
\centering
%\resizebox{1\textwidth}{!}{
\begin{tabular}{cccc}
\toprule
%\cmidrule(lr){3-5} \cmidrule(lr){6-7}
\textbf{Model} & \textbf{$1$-st module} & \textbf{$2$-nd module} & \textbf{Total} \\
\midrule
NestAR-B&172M&172M&344M\\
NestAR-L&172M&608M&780M\\
NestAR-H&172M&1.1B&1.3B\\ 
\bottomrule
\end{tabular}
%}
\caption{AutoRegressive model sizes for Basic, Large and Huge variants.}
\label{tab:size}
\end{table}

\begin{figure}[h]
    \includegraphics[scale=0.22]{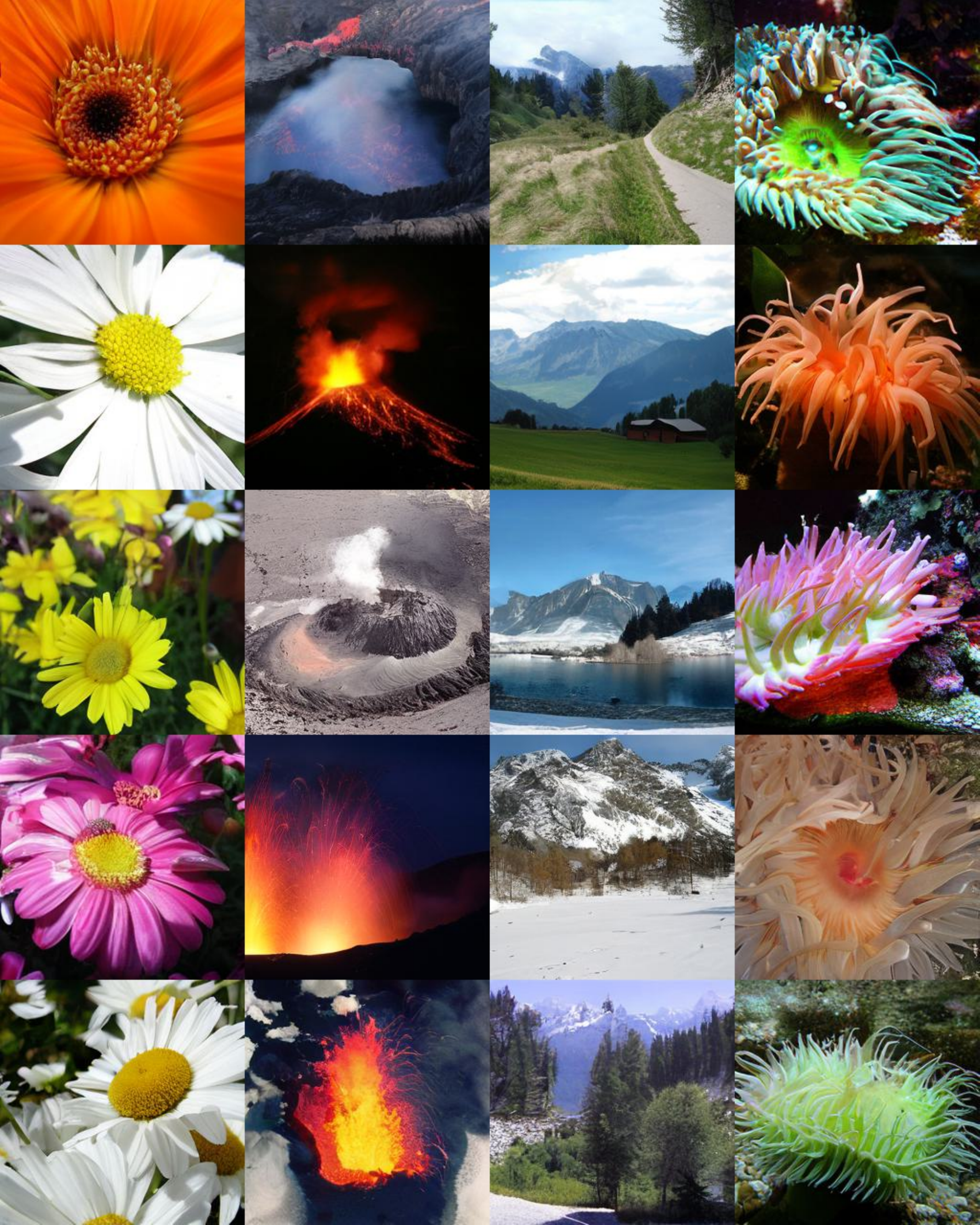}
    \centering
    \caption{Qualitative Results: $256\times 256$ image samples of the same classes to demonstrate diversity of images. The classes from left to right are: Daisy, Volcano, Alps, and Coral. }
    \label{image2}
\end{figure}

\subsection{Main Results - RQ1}
We evaluate the performance of NestAR using FID \cite{fid}, Inception Score IS \cite{is}, Precision, and Recall. FID measures the similarity between real and generated image distributions by comparing their feature embeddings, with lower values indicating higher fidelity and diversity. IS assesses how well the generated images resemble distinct, meaningful objects. Higher IS suggests more realistic and varied images. Table~\ref{tab:main} compares NestAR's performance with state-of-the-art generative models.

Our best variant of NestAR achieves an FID of $2.22$ and an IS of $342.4$. While the FID is  not the highest compared to other continuous-token models such as MAR~\cite{mar} and xAR~\cite{xar}, its performance is comparable to discrete-token models such as LlamaGen~\cite{llamagen}, VAR\cite{var} and MonoFormer~\cite{mono}, and better than others such as ViTVQ\cite{vit} and DART-AR\cite{dart}. NestAR achieves the  best IS score of $342.4$, beating the previous SOTA score made by VAR~\cite{var} by $5.9\%$. The exceptional IS score demonstrates NestAR's success in boosting the diversity of generated images.  

\subsection{Generation Speed - RQ2}
In this section, we compare the generation speed of NestAR with other popular image generative models in Table~\ref{tab:speed}. Our smallest and largest variants, NestAR-B and NestAR-H, are nearly 20 times and 3 times faster than MAR\cite{mar}, diffusion, and flow matching models \cite{repa,dit,sit} respectively. NestAR is only slightly slower than xAR \cite{xar}. Qualitatively, this is due to the additional scaled modules in NestAR. However,  this has not had a substantial impact on the generation speed because the additional scaled modules generate significantly smaller tokens, leading to faster speed. In addition, the larger scaled module now has one less token to generate. These factors intuitively lead to NestAR having a slightly reduced speed, though it remains largely comparable.

% %\begin{table}[!htbp]
% \begin{table}[h]
% \centering
% %\resizebox{1\textwidth}{!}
% {\footnotesize
% \begin{tabular}{cccccc}
% \toprule
% %\cmidrule(lr){3-5} \cmidrule(lr){6-7}
% \textbf{Type} & \textbf{Model} & \textbf{Params}  & \textbf{FID}$\downarrow$ & \textbf{steps} & \textbf{img/sec} \\
% \midrule
% Diff.&DiT-XL/2&675M&2.27&250&0.5\\
% \midrule 
% FM&SiT-XL/2&675M&2.06&250&0.5\\
% FM&REPA&675M&1.8&250&0.6\\
% \midrule 
% MAR&MAR-L&479M&1.78&256&0.5\\
% MAR&MAR-H&943M&1.55&256&0.3\\
% \midrule 
% xAR&XAR-B&172M&1.72&50&9.8\\
% xAR&XAR-L&608M&1.28&50&3.2\\
% xAR&XAR-H&1.1B&1.24&50&1.3\\
% \midrule 
% NestAR&NestAR-B&344M&2.86&50&9.5\\
% NestAR&NestAR-L&780M&2.29&50&3.5\\
% NestAR&NestAR-H&1.3B&2.22&50&1.6\\
% \midrule 
% \bottomrule
% \end{tabular}
% }
% \caption{Comparison of generation speed on ImageNet-256. Throughput for NestAR is evaluated as images generated per second on a single A100. Other models' metrics are from \cite{xar}. \xuhui{looks like xAR is better in every aspect. need to emphasize the unique advantages of NestAR.}\hongyu{should I take out the FID column, since this is a speed comparison table?}}
% \label{tab:speed}
% \end{table}

\begin{table}[h]
\centering
%\resizebox{1\textwidth}{!}
{\footnotesize
\begin{tabular}{ccccc}
\toprule
%\cmidrule(lr){3-5} \cmidrule(lr){6-7}
\textbf{Type} & \textbf{Model} & \textbf{Params} & \textbf{steps} & \textbf{img/sec} \\
\midrule
Diff.&DiT-XL/2&675M&250&0.5\\
\midrule 
FM&SiT-XL/2&675M&250&0.5\\
FM&REPA&675M&250&0.6\\
\midrule 
MAR&MAR-L&479M&256&0.5\\
MAR&MAR-H&943M&256&0.3\\
\midrule 
xAR&XAR-B&172M&50&9.8\\
xAR&XAR-L&608M&50&3.2\\
xAR&XAR-H&1.1B&50&1.3\\
\midrule 
NestAR&NestAR-B&344M&50&9.5\\
NestAR&NestAR-L&780M&50&3.5\\
NestAR&NestAR-H&1.3B&50&1.6\\
\midrule 
\bottomrule
\end{tabular}
}
\caption{Comparison of generation speed on ImageNet-256. Throughput for NestAR is evaluated as images generated per second on a single A100. The metrics of other models are from \cite{xar}. }
\label{tab:speed}
\end{table}

\subsection{Qualitative Results}
NestAR is capable of producing photo quality images with high fidelity. We present images generated from NestAR-H at the $256 \times 256$ resolution in Figure~\ref{image1}.

Figure~\ref{image2} illustrates sampled images from the same classes to demonstrate the diverse range of images that NestAR is capable of generating. As we can see, given the same class, the proposed NestAR is able to generate highly diversified images. Take Daisy and Coral as examples, the generated images can be in different colors and different shapes.

\subsection{Sensitivity to sizes of the $1$-st module - RQ3}
\begin{table}[h]
%\begin{table}[!htbp]
\centering
%\resizebox{1\textwidth}{!}{
\begin{tabular}{ccccc}
\toprule
%\cmidrule(lr){3-5} \cmidrule(lr){6-7}
\textbf{$1$-st module} & \textbf{$2$-nd module} & \textbf{Total}  & \textbf{FID}$\downarrow$ & \textbf{IS}$\uparrow$ \\
\midrule
33M&172M&205M&2.97&263.3\\
58M&172M&230M&2.91&285.3\\
172M&172M&344M&2.86&320.6\\
\bottomrule
\end{tabular}
%}
\caption{Sensitivity to the size of $1$-st module.}
\label{tab:ab1}
\end{table}
In this section, we study the effect of the size of the first scaled module in a two level NestAR on the modeling performance. We hold the second layer from NestAR-B constant and use different sizes of first layer models to measure the performance. The result is shown in Table~\ref{tab:ab1}.

The result indicates that the size of $1$-st scaled module does not have a significant impact on FID, reducing the size by 5 times, the FID increases by ~4\%, and IS decreases by ~17\%. Intuitively, the $1$-st scaled module has a smaller number of features to learn. Hence, a smaller model would be sufficient, increasing its size does not achieve much better results.

\section{Conclusions}
A Nested AutoRegressive model NestAR is proposed to address the inefficiency and diversity limitations of existing AutoRegressive image generation approaches. NestAR reduces the generation complexity from $\mathcal{O}(n)$ to $\mathcal{O}(\log n)$ through a novel two-level AutoRegressive design and improves sample diversity by modeling within-scale dependencies. By incorporating continuous tokens and flow-matching loss, our model generates high-quality images with enhanced efficiency. Experimental results demonstrate that NestAR achieves competitive performance against state-of-the-art models while offering significant improvements in inference speed and sample diversity.

% \clearpage
\bibliography{aaai26-NesAR}

\end{document}